\documentclass[runningheads]{llncs}
\usepackage{graphicx}
\usepackage{marvosym}
\usepackage{amsmath,amssymb,amsfonts}
\usepackage{algorithmic}
\usepackage{graphicx}
\usepackage{textcomp}
\usepackage[T5]{fontenc}
\usepackage[utf8]{inputenc}
\usepackage[vietnamese,english]{babel}
\usepackage{pifont}
\usepackage{float}
\usepackage{caption}
\usepackage{placeins}
\usepackage{array}
\newcolumntype{P}[1]{>{\centering\arraybackslash}p{#1}}
\newcolumntype{M}[1]{>{\centering\arraybackslash}m{#1}}
\usepackage{multirow}
\usepackage{hyperref}
\usepackage{amsmath}
\hypersetup{colorlinks, citecolor=blue, linkcolor=blue, urlcolor=blue}

\begin{document}
\title{Emotion Recognition\\for Vietnamese Social Media Text}

\titlerunning{Emotion Recognition for Vietnamese Social Media Text}

\author{Vong Anh Ho\inst{1,4}\textsuperscript{(\Letter)} \and
Duong Huynh-Cong Nguyen\inst{1,4} \and
Danh Hoang Nguyen\inst{1,4} \and
\\Linh Thi-Van Pham\inst{2,4} \and
Duc-Vu Nguyen\inst{3,4} \and
\\Kiet Van Nguyen\inst{1,4} \and
Ngan Luu-Thuy Nguyen\inst{1,4}}

\authorrunning{Vong Anh Ho et al.}

\institute{University of Information Technology, VNU-HCM, Vietnam\\
\email{\{15521025,15520148,15520090\}@gm.uit.edu.vn, \{kietnv,ngannlt\}@uit.edu.vn}\\
\and
University of Social Sciences and Humanities, VNU-HCM, Vietnam\\
\email{vanlinhpham888@gmail.com}\\
\and
Multimedia Communications Laboratory, University of Information Technology, VNU-HCM, Vietnam\\
\email{vund@uit.edu.vn}\\
\and
Vietnam National University, Ho Chi Minh City, Vietnam}
\maketitle

\begin{abstract}

Emotion recognition or emotion prediction is a higher approach or a special case of sentiment analysis. In this task, the result is not produced in terms of either polarity: positive or negative or in the form of rating (from 1 to 5) but of a more detailed level of analysis in which the results are depicted in more expressions like sadness, enjoyment, anger, disgust, fear and surprise. Emotion recognition plays a critical role in measuring brand value of a product by recognizing specific emotions of customers' comments. In this study, we have achieved two targets. First and foremost, we built a standard \textbf{V}ietnamese \textbf{S}ocial \textbf{M}edia \textbf{E}motion \textbf{C}orpus (UIT-VSMEC) with exactly 6,927 emotion-annotated sentences, contributing to emotion recognition research in Vietnamese which is a low-resource language in natural language processing (NLP). Secondly, we assessed and measured machine learning and deep neural network models on our UIT-VSMEC corpus. As a result, the CNN model achieved the highest performance with the weighted F1-score of 59.74\%. Our corpus is available at our research website \footnote[1]{\url{ https://sites.google.com/uit.edu.vn/uit-nlp/corpora-projects}}.

\end{abstract}

\section{Introduction} \label{introduction}
Expressing emotion is a fundamental need of human and that we use language not just to convey facts, but also our emotions \cite{Kiritchenko}. Emotions determine the quality of our lives, and we organize our lives to maximize the experience of positive emotions and minimize the experience of negative emotions \cite{Ekman}. Thus, Paul Ekman \cite{Ekman1993} proposed six basic emotions of human including enjoyment, sadness, anger, surprise, fear, and disgust through facial expression. Nonetheless, apart from facial expression, many different sources of information can be used to analyze emotions since emotion recognition has emerged as an important research area. And in recent years, emotion recognition in text has become more popular due to its vast potential applications in marketing, security, psychology, human-computer interaction, artificial intelligence, and so on \cite{Mohammad}.

In this study, we focus on the problem of recognizing emotions for Vietnamese comments on social network. To be more specific, the input of the problem is a Vietnamese comment from social network, and the output is a predicted emotion of that comment labeled with one of these: enjoyment, sadness, anger, surprise, fear, disgust and other. Several examples are shown in Table~\ref{tab:1}.

\begin{table}[!htbp]
\caption{Examples of emotion-labeled sentences.}
\label{tab:1}
\vspace{10pt}
\begin{tabular}{|c|p{5cm}|p{4.5cm}|c|}
\hline
\textbf{No.} & \multicolumn{1}{c|}{\textbf{Vietnamese sentences}}           & \multicolumn{1}{c|}{\textbf{English translation}}          & \textbf{Emotion}   \\ \hline
1   & Ảnh đẹp quá!                                        & The picture is so beautiful!                      & Enjoyment \\ \hline
2   & Tao khóc..huhu.. Tao rớt rồi                        & I'm crying..huhu.. I failed the exam.             & Sadness   \\ \hline
3   & Khuôn mặt của tên đó vẫn còn ám ảnh tao.             & The face of that man still haunts me.             & Fear      \\ \hline
4   & Cái gì cơ? Bắt bỏ tù lũ khốn đó hết!                & What the fuck? Arrest all those goddamn bastards! & Anger     \\ \hline
5   & Thật không thể tin nổi, tại sao lại nhanh đến thế?? & It's unbelievable, why can be that fast??         & Surprise  \\ \hline
6   & Những điều nó nói làm tao buồn nôn.                  & What he said makes me puke.                       & Disgust   \\ \hline
\end{tabular}
\end{table}

In this paper, our two key contributions are summarized as follows.
\begin{itemize}
\item One of the most primary contributions is to obtain the UIT-VSMEC corpus, which is the first corpus for emotion recognition for Vietnamese social media text. As a result, we achieved 6,927 emotion annotated-sentences. To ensure that only the best results with high consistency and accuracy are reached, we built a very coherent and thorough annotation guideline for the dataset. The corpus is publicly available for research purpose.

\item The second one, we tried using four learning algorithms on our UIT-VSMEC corpus, two machine learning models consisting of Support Vector Machine (SVM) and Random Forest versus two deep learning models including Convolutional Neural Network (CNN) and Long Short-Term Memory (LSTM).
\end{itemize}

The structure of the paper is organized thusly. Related documents and studies are presented in Section II. The process of building corpus, annotation guidelines, and dataset evaluation are described in Section III. In Section IV, we show how to apply SVM, Random Forest, CNN, and LSTM for this task. The experimental results are analyzed in Section V. Conclusion and future work are deduced in Section VI.
\section{Related Work} \label{wsmethod}
There are some related work in English and Chinese. In 2007, the SemEval-2007 Task 14 \cite{CarloStrapparava}  developed a dataset for emotion recognition with six emotion classes (enjoyment, anger, disgust, sadness, fear and surprise) including 1,250 newspaper headline human-annotated sentences. In 2012, Mohammad \cite{Mohammad} published an emotion corpus with 21,052 comments from Tweets annotated also by six labels of emotion (enjoyment, anger, disgust, sadness, fear and surprise). In 2017, Mohammad \cite{SaifMohammad2017} again published a corpus annotated with only four emotion labels (anger, fear, enjoyment and sadness) for 7,079 comments from Tweets. In 2018, Wang \cite{ZhongqingWang} put out a bilingual corpus in Chinese and English for emotion recognition including 6,382 sentences tagged by five different emotions (enjoyment, sadness, fear, anger and surprise). In general, corpus for emotion recognition task use some out of six basic emotions of human (enjoyment, sadness, anger, disgust, fear and surprise) based on Ekman's emotion theory \cite{Ekman1993}.

In terms of algorithms, Kratzwald \cite{BernhardKratzwald} tested the efficiency of machine learning algorithms (Random Forest and SVM) and deep learning algorithms (Long Short-Term Memory (LSTM) and Bidirectional Long Short-Term Memory (BiLSTM)) combined with pre-trained word embeddings on multiple emotion corpora. In addition, the BiLSTM combined with pre-trained word embeddings reached the highest result of 58.2\% of F1-score compared to Random Forest and SVM with 52.6\% and 54.2\% respectively on the General Tweets corpus \cite{BernhardKratzwald}. Likewise, Wang \cite{TingweiWang} proposed the Bidirectional Long Short-Term Memory Multiple Classifiers (BLSTM-MC) model on a bilingual corpus in Chinese and English that achieved the F1-score of 46.7\%, ranked third in the shared task NLPCC2018 - Task 1 \cite{ZhongqingWang}.

In Vietnamese, there are quite a quantity of research works on other NLP tasks such as parsing \cite{nguyen2014treebank,nguyen2016vietnamese}, part-of-speech \cite{bach2018empirical,nguyen2017word}, named entity recognition \cite{thao2007named,nguyen2016approach}, sentiment analysis \cite{KietVanNguyen,PhuNguyen,VuDucNguyen} and question answering \cite{Nguyen_2009,le2018factoid}.  However, there are no research publications on emotion recognition for Vietnamese social media texts.  Therefore, we decided to build a first corpus of Vietnamese emotion recognition for the research purposes. In this paper, we evaluate machine learning (SVM and Random Forest) and deep learning models (CNN and LSTM) on our corpus.


\section{Corpus Construction}
In this section, we present the process of developing the UIT-VSMEC corpus in Section 3.1, annotation guidelines in Section 3.2, corpus evaluation in Section 3.3 and corpus analysis in Section 3.4.
\subsection{Process of Building the Corpus}

\begin{figure}[!htbp]
\centering
\includegraphics[width=\textwidth]{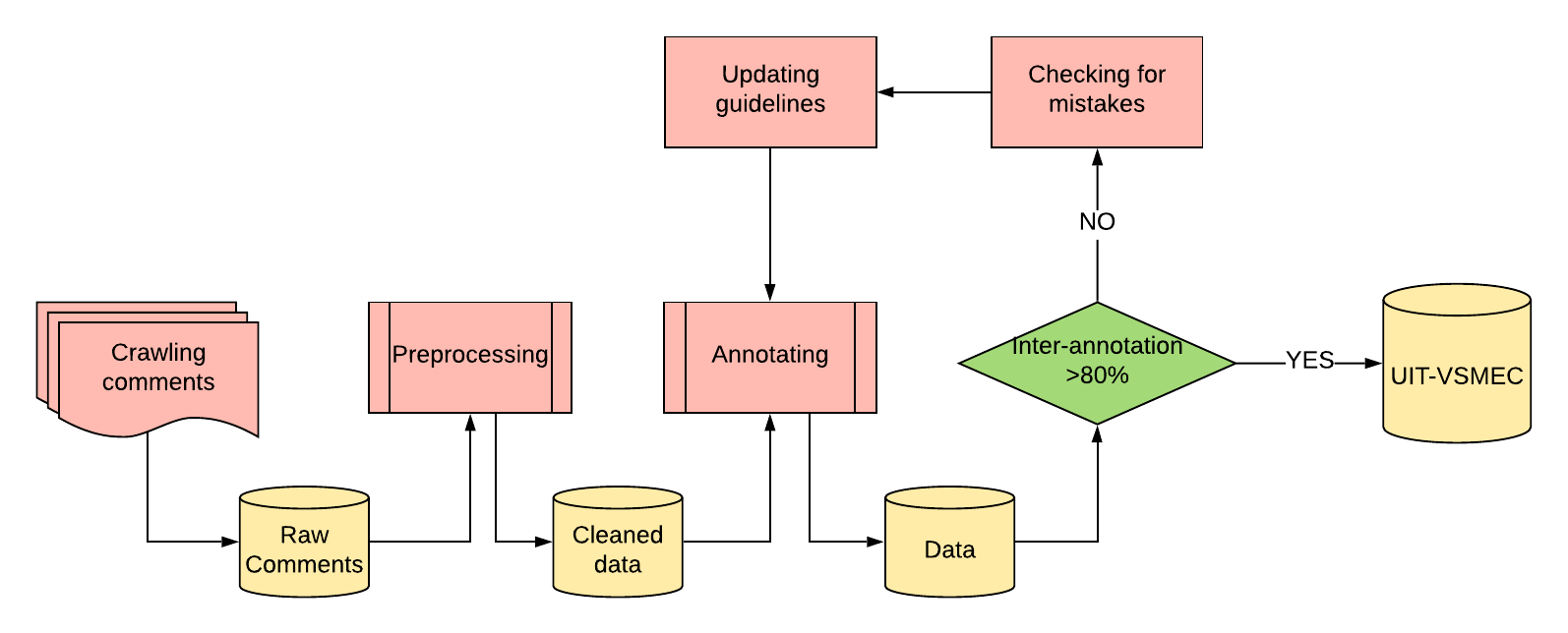}
\caption{Overview of corpus-building process.}
\label{fig1}
\end{figure}

The overview of corpus-building process which includes three phases is shown in Figure \ref{fig1} and the detailed description of each phase is presented shortly thereafter.

\textbf{Phase 1 - Collecting data}: We collect data from Facebook which is the most popular social network in Vietnam. In particular, according to a survey of 810 Vietnamese people conducted by \cite{Jointstockcompany} in 2018, Facebook is used the most in Vietnam. Moreover, as reported by the statistic of the number of Facebook users by \cite{Nguyen}, there are 58 million Facebook users  in Vietnam, ranking as the 7th country with the most users in the world. As the larger the number of the users and the more interaction between them, the richer the data. And to collect the data, we use Facebook API to get Vietnamese comments from public posts.

\textbf{Phase 2 - Pre-processing data}: To ensure users' privacy, we replace users' names in the comments with PER tag. The rest of the comments is kept as it is to retain the properties of a comment on social network.

\textbf{Phase 3 - Annotating data}: This step is divided into two stages: \textbf{Stage 1}, building annotation guidelines and training for four annotators. Data tagging is repeated for 806 sentences along with editing guidelines until the consensus reaches more than 80\%. \textbf{Stage 2}, 6,121 sentences are shared equally for three annotators while the another checking the whole 6,121 sentences and the consensus between the two is above 80\%.

\subsection{Annotation Guidelines}
There are many suggestions on the number of basic emotions of human. According to studies in the field of psychology, there are two outstanding viewpoints in basic human emotions: "Basic Emotions" by Paul Ekman and "Wheel of Emotions" by Robert Plutchik \cite{Kiritchenko}. Paul Ekman's studies point out that there are six basic emotions which are expressed through the face \cite{Ekman1993}: enjoyment, sadness, anger, fear, disgust and surprise. Eight years later, Robert Plutchik gave an another definition of emotion. In this concept, emotion is divided into eight basic ones which are polarized in pairs: enjoyment - sadness, anger - fear, trust - disgust and surprise - anticipation \cite{Kiritchenko}. Despite the agreement on basic emotions, psychologists dissent from the number of which are the most basic that may have in humans, some ideas are 6, 8 and 20 or even more \cite{SaifMohammad2017}.

To that end, we chose six labels of basic emotions (enjoyment, sadness, anger, fear, disgust and surprise) for our UIT-VSMEC corpus based on six basic human emotions proposed by Ekman \cite{Ekman1993} together with Other label to mark a sentence with an emotion out of the six above or a sentence without emotion, considering most of the automated emotion recognition work in English \cite{SaifMohammad,Kiritchenko} are all established from Ekman's emotion theory (1993) and with a large amount of comments in the corpus, a small number of emotions makes the manual tagging process more convenient.

Based on Ekman's instruction in basic human emotions \cite{PaulEkman}, we build annotation guidelines for Vietnamese text with seven emotion labels described as follows.
\begin{itemize}
\item \textbf{Enjoyment}: For comments with the states that are triggered by feeling connection or sensory pleasure. It contains both peace and ecstasy. The intensity of these states varies from the enjoyment of helping others, a warm uplifting feeling that people experience when they see kindness and compassion, an experience of ease and contentment or even the enjoyment of the misfortunes of another person to the joyful pride in the accomplishments or the experience of something that is very beautiful and amazing. For example, the emotion of the sentence "Nháy mắt thôi cũng đáng yêu, kkk" (English translation: "Just the act of winking is so lovely!") is Enjoyment.

\item \textbf{Sadness}: For comments that contain both disappointment and despair. The intensity of its states varies from discouragement, distraughtness, helplessness, hopelessness to strong suffering, a feeling of distress and sadness often caused by a loss or sorrow and anguish. The Vietnamese sentence "Lúc đấy khổ lắm... kỉ niệm :(" (English translation: "It was hard that time..memory :(" ) has an emotion of Sadness, for instance.

\item \textbf{Fear}: For comments that show anxiety and terror. The intensity of these states varies from trepidation - anticipation of the possibility of danger, nervousness, dread to desperation, a response to the inability to reduce danger, panic and horror - a mixture of fear, disgust and shock. A given sentence "Chuyện này làm tao nổi hết da gà" (English translation: "This story causes me goosebumps") is a Fear labeled-sentence.

\item \textbf{Anger}: For comments with states that are triggered by a feeling of being blocked in our progress. It contains both annoyance and fury and varies from frustration which is a response to repeated failures to overcome an obstacle, exasperation - anger caused by strong nuisance, argumentativeness to bitterness - anger after unfair treatment and vengefulness. For example, "Biến mẹ mày đi!" (English translation: "You fucking get lost!") is labeled with Angry.

\item \textbf{Disgust}: For comments which show both dislike and loathing. Their intensity varies from an impulse to avoid something disgusting or aversion, the reaction to a bad taste, smell, thing or idea, repugnance to revulsion which is a mixture of disgust and loathing or abhorrence - a mixture of intense disgust and hatred. As "Làm bạn với mấy thể loại này nhục cả người" (English translation: "Making friends with such types humiliates you") has an emotion of Disgust.

\item \textbf{Surprise}: For comments that express the feeling caused by unexpected events, something hard to believe and may shock you. This is the shortest emotion of all emotions, only takes a few seconds. And it passes when we understand what is happening, and it may become fear, anger, relief or nothing ... depends on the event that makes us surprise. "Trên đời còn tồn tại thứ này sao??" (English translation: "How the hell in this world this still exists??") is annotated with Surprise.

\item \textbf{Other}: For comments that show none of those emotions above or comments that do not contain any emotions. For instance, "Mình đã xem rất nhiều video như này rồi nên thấy cũng bình thường" (English translation: "I have seen a lot of videos like this so it's kinda normal") is neutral, so its label is Other.
\end{itemize}
\subsection{Corpus Evaluation}

We use the A\textit{\textsubscript{m}} agreement measure  \cite{PlabanKumarBhowmick} to evaluate the consensus between annotators of the corpus. This agreement measure was also utilized in the UIT-VSFC corpus \cite{KietVanNguyen}. A\textit{\textsubscript{m}}  is calculated by the following formula.
\[A_m = \frac{P_o - P_e}{1 - P_e}\]

Where, P\textit{\textsubscript{o}} is the observed agreement which is the proportion of sentences with both of the annotators agreed on the classes pairs and P\textit{\textsubscript{e}} is the expected agreement that is the proportion of items for which agreement is expected by chance when the sentences are seen randomly. 

Table \ref{tab:2} presents the consensus of the entire UIT-VSMEC corpus with two separate parts. The consensus of 806 sentences which can be seen as the first stage of annotating data mentioned in section 3-3.1. A\textit{\textsubscript{m}} agreement level is high with 82.94\% by four annotators. And the annotation agreement in the second stage (3-3.1) where X$_1$, X$_2$, X$_3$ are 3 annotators  independently tagging data and Y is the checker of one another. The A\textit{\textsubscript{m}} and P\textit{\textsubscript{o}} of annotation pair X$_3$-Y are highest of 89.12\% and 92.81\%.

\begin{table}[]
\centering
\caption{Annotation agreement of the UIT-VSMEC corpus (\%)}
\label{tab:2}
\begin{tabular}{|c|c|c|c|c|}
\hline
\textbf{Stage} & \textbf{Annotators} & \textbf{P$_0$} & \textbf{P$_e$} & \textbf{A$_m$} \\ \hline
\multicolumn{1}{|l|}{1} & X$_1$-X$_2$-X$_3$-Y (806 sentences) & 92.66 & 56.98 & 82.94 \\ \hline
\multicolumn{1}{|l|}{2} & X$_1$-Y (2,032 sentences) & 88.00 & 18.61 & 85.25 \\ 
\multicolumn{1}{|l|}{} & X$_2$-Y (2,112 sentences) & 86.27 & 31.23 & 80.03 \\ 
\multicolumn{1}{|l|}{} & X$_3$-Y (1,977 sentences) & 92.81 & 33.95 & 89.12 \\ \hline
\end{tabular}
\end{table}

\subsection{Corpus Analysis}
After building the UIT-VSMEC corpus, we obtained 6,927 human-annotated sentences with one of the seven emotion labels. Statistics of emotion labels of the corpus is presented in Table \ref{tab:3}.

\begin{table}[!h]
\centering
\caption{Statistics of emotion labels of the UIT-VSMEC corpus }
\label{tab:3}
\begin{tabular}{|l|r|r|}
\hline
\textbf{Emotion} & \textbf{Sentences} & \textbf{Percentage} (\%) \\ \hline
Enjoyment & 1,965 & 28.36 \\ \hline
Disgust & 1,338 & 19.31 \\ \hline
Sadness & 1,149 & 16.59 \\ \hline
Anger & 480 & 6.92 \\ \hline
Fear & 395 & 5.70 \\ \hline
Surprise & 309 & 4.46 \\ \hline
Other & 1,291 & 18.66 \\ \hline
Total & 6,927 & 100 \\ \hline
\end{tabular}
\end{table}


Through Table \ref{tab:3}, we concluded that the comments got from social network are uneven in number among different labels in which the enjoyment label reaches the highest number of 1,965 sentences (28.36\%) while the surprise label arrives at the lowest number of 309 sentences (4.46\%).

Besides, we listed the number of the sentences of each label up to their lengths. Table \ref{tab:4} shows the distribution of emotion-annotated sentences according to their lengths. It is easy to see that most of the comments are from 1 to 20 words accounting for 81.76\%.

\begin{table}[h]
\centering
\caption{Distribution of emotion-annotated sentences according to the length of the sentence (\%)}
\label{tab:4}
\setlength\tabcolsep{2.1pt}
\begin{tabular}{|c|c|c|c|c|c|c|c|c|}
\hline
\textbf{Length} & \textbf{Enjoyment} & \textbf{Disgust} & \textbf{Sadness} & \textbf{Anger} & \textbf{Fear} & \textbf{Surprise} & \textbf{Other} & \textbf{Overall} \\ \hline
1-5 & 5.16 & 2.84 & 1.70 & 0.69 & 0.94 & 0.85 & 1.87 & \textbf{14.05} \\ \hline
6-10 & 8.98 & 4.22 & 4.98 & 1.41 & 1.42 & 2.25 & 7.20 & \textbf{30.38} \\ \hline
10-15 & 5.87 & 3.99 & 4.11 & 1.40 & 1.27 & 0.94 & 5.00 & \textbf{22.58} \\ \hline
16-20 & 4.17 & 3.05 & 2.51 & 1.14 & 0.85 & 0.24 & 2.79 & \textbf{14.75} \\ \hline
21-25 & 1.96 & 1.93 & 1.50 & 0.66 & 0.40 & 0.15 & 1.11 & 7.71 \\ \hline
26-30 & 1.08 & 1.31 & 0.95 & 0.45 & 0.27 & 0.01 & 0.53 & 4.6 \\ \hline
\textgreater{}30 & 1.23 & 1.97 & 0.84 & 1.17 & 0.55 & 0.02 & 0.15 & 5.93 \\ \hline
Total & 28.36 & 19.31 & 16.59 & 6.92 & 5.70 & 4.46 & 18.66 & 100 \\ \hline
\end{tabular}
\end{table}

\section{Methodology}
In this paper, we use two kinds of methodologies to evaluate the UIT-VSMEC corpus including two machine learning models (Random Forest and SVM) and two deep learning models (CNN and LSTM) as the first models described as follows.
\subsection{Machine Learning Models}
The authors in \cite{BernhardKratzwald} proposed SVM and Random Forest algorithms for emotion recognition. Under which, we also tested three more machine learning algorithms including Decision Tree, kNN and Naive Bayes on 1,000 emotion-annotated sentences extracted from the UIT-VSMEC corpus by Orange3. Consequently, Random Forest achieved the second best result after SVM which is displayed in Table \ref{tab:5}. It is the main reason why we chose SVM and Random Forest for experiments on the UIT-VSMEC corpus.

\begin{table}[h]
\centering
\caption{Experimental results by Orange3 of machine learning models on 1,000 emotion-annotated sentences from the UIT-VSMEC corpus (\%)}
\label{tab:5}
\begin{tabular}{|l|l|l|}
\hline
\textbf{Method}        & \textbf{Accuracy} & \textbf{Weighted F1} \\ \hline
\textbf{Random Forest} & 35.8              & \textbf{32.8}        \\ \hline
\textbf{SVM}           & 37.6              & \textbf{37.0}        \\ \hline
Decision Tree          & 30.5              & 29.6                 \\ \hline
kNN                    & 28.9              & 27.1                 \\ \hline
Naĩve Bayes            & 20.8              & 19.2                 \\ \hline
\end{tabular}
\end{table}

\subsubsection{Random Forest}
Random Forest is a versatile machine learning algorithm when used for classification problems, predicting linear regression values and multi-output tasks. The idea of Random Forest is to use a set of Decision Tree classification, each of which is trained on different parts of the dataset. After that, Random Forest will get back all the classification results of the seedlings from which it chooses the most voted one to give the final result. Despite of its simplicity, Random Forest is one of the most effective machine learning algorithms today \cite{AurelienGeron}. 
\subsubsection{Support Vector Machine (SVM)}
 We use the SVM machine learning algorithm as a baseline result for this emotion recognition problem. According to the authors in \cite{SaifMohammad}, SVM is an effective algorithm for classification problems with high features. Here, we use SVM model supported by scikit-learn library. 
\subsection{Deep Learning Models}
\subsubsection{Long Short-Term Memory (LSTM)}
LSTM is also applied for the UIT-VSMEC corpus for various reasons. To begin with, LSTM is considered as the state-of-the-art method of almost sequence prediction problems. Moreover, through the two competitions WASSA-2018 \cite{RomanKlinger} and SemEval-2018 Task 1 \cite{Mohammad2018} for emotion recognition task, we acknowledged that LSTM was effectively used the most. Furthermore, LSTM has advantages over conventional neural networks and Recurrent Neural Network (RNN) in many different ways due to its selective memory characteristic in a long period. This is also the reason why the authors in \cite{BernhardKratzwald} chose to use it in his paper. Therefore, we decided to use LSTM on the same problem on our corpus.

LSTM consists of four main parts: Word embeddings input, LSTM cell network, fully connected and softmax. With the input, each cell in the LSTM network receives a word vector represented by word embeddings with the form [1 x n] where n is the fixed length of the sentence. Then cells calculate the values and gets the results as vectors in LSTM cell network. These vectors will go through fully connected and the output values will then pass through softmax function to give an appropriate classification for each label.

\subsubsection{Convolutional Neural Network (CNN)}
 We use Convolutional Neural Network (CNN) algorithm which is proposed in \cite{Kim} to recognize emotions in a sentence. CNN is the algorithm that achieves the best results in four out of the seven major problems of Natural Language Processing which includes both emotion recognition and question classification tasks (text classification, language model, speech recognition, title generator, machine translation, text summarization and Q\&A systems) \cite{Kim,YingjieZhang}.

A CNN model consists of three main parts: Convolution layer, pooling layer and fully connected layer. In convolution layer - the Kernel, we used 3 types of filters of different sizes with total 512 filters to extract the high-level features and obtain convolved feature maps. These then go through the pooling layer which is responsible for reducing the spatial size of the convolved feature and decreasing the computational power required to process the data through dimensionality reduction. The convolutional layer and the pooling layer together form the i-th layer of a Convolutional Neural Network. Moving on, the final output will be flattened and fed to a regular neural network in the fully connected layer for classification purposes using the softmax classification technique.





\section{Experiments and Error Analysis} \label{experiment and Evaluation}
\subsection{Corpus Preparation}

We at first built a normalized corpus for comparison and evaluation where spelling errors have been corrected and acronyms in various forms have been converted back to their original words, seeing it is impossible to avoid such problems of text on social networks when it does not distinguish any type of users. Table \ref{tab:6} shows some examples being encountered the most in the dataset.

\begin{table}[!htp]
\centering
\caption{Vietnamese abbreviations in the dataset.}
\label{tab:6}
\begin{tabular}{|c|l|l|l|}
\hline
\textbf{No.} & \multicolumn{1}{c|}{\textbf{Abbreviation}}& \multicolumn{1}{c|}{\textbf{Vietnamese meaning}}& \multicolumn{1}{c|}{\textbf{English meaning}}\\ \hline
1 & “dc” or “dk” or  “duoc” & "được" & "ok" \\ \hline
2 & “ng” or “ngừi” & "người" & "people" \\ \hline
3 & "trc" or "trk" & "trước" & "before" \\ \hline
4 & "cg" or "cug" or "cũg" & "cũng" & "also" \\ \hline
5 & "mk" or "mik" or "mh" & "mình" & "I" \\ \hline
\end{tabular}
\end{table}

We then divided the UIT-VSMEC corpus into the ratio of 80:10:10, in which 80\% of the corpus is the training set, 10\% is the validation one and the rest is the test set. The UIT-VSMEC corpus is an imbalanced-labels corpus, therefore, to ensure that sentences in low-volume labels are distributed fully in each set, we use stratified sampling method utilizing train\_test\_split() function supported by scikit learn library to distribute them into training, validation and test sets. The result is presented in Table \ref{tab:7}. 
\begin{table}[!htp]
\centering
\caption{Statistics of emotion-labeled sentences in training, validation and test sets.}
\label{tab:7}
\begin{tabular}{|l|r|r|r|r|}
\hline 
 \textbf{Emotion} & \textbf{Train} & \textbf{Dev} & \textbf{Test} & \textbf{Total} \\ \hline
Enjoyment & 1,573 & 205 & 187 & 1,965 \\ \hline
Disgust & 1,064 & 141 & 133 & 1,338 \\ \hline
Sadness & 938 & 92 & 119 & 1,149 \\ \hline
Anger & 395 & 38 & 47 & 480 \\ \hline
Fear & 317 & 38 & 47 & 395 \\ \hline
Surprise & 242 & 36 & 31 & 309 \\ \hline
Other & 1,019 & 132 & 140 & 1,291 \\ \hline
All & 5,548 & 686 & 693 & 6,927 \\ \hline
\end{tabular}
\end{table}

\subsection{Experimental Settings}

In this paper, to represent words in vector form, we use two different methods word embeddings and bag of words. For the two machine learning models SVM and Random Forest, we use bag of words in conjunction with TF-IDF. For the two other deep learning models LSTM and CNN, we utilize pre-trained word embeddings including word2vec \footnote[3]{\url{ https://github.com/vncorenlp/VnCoreNLP}} and fastText \footnote[4]{\url{https://fasttext.cc/docs/en/crawl-vectors.html}} used as its main techniques. 

With machine learning models SVM and Random Forest, grid-search method is utilized to get the most appropriate parameters for the task. In particular, with SVM we use word of tag (1, 3) combined with bag of char (1,7) features and loss function hinge, and to reduce overfitting we apply the l2-regularization technique with lambd. = 1e-4. About Random Forest model, the number of decision trees is 256 and the depth of the trees is 64.

For LSTM model, we use the many-to-one architecture due to the classification requirement of the problem. To select proper parameters for emotion recognition in Vietnamese, we add two drop-out classes of 0.75 and 0.5 respectively to increase processing time as well as to avoid overfitting.

Regarding deep learning CNN model, we apply three main kernels: 3, 4 and 5 with a number of each is 128. Besides, drop-out of 0.95 and l2 of 0.01 are adopted to avoid overfitting. Properties and models are developed from Yoon Kim's work \cite{Kim}.




\subsection{Experimental Results}

In this section, we present the results of two experiments. Firstly, we test and compare the results of each model on the UIT-VSMEC corpus. Secondly, we evaluate the influence of Other label on this corpus after implementing these machine learning and deep learning models on the corpus yet without Other label. All models are evaluated by accuracy and weighted F1-score metrics.
\begin{table}[htb]
\centering
\caption{Experimental results of the UIT-VSMEC corpus.}
\label{tab:8}
\begin{tabular}{|l|l|r|r|}
\hline
\textbf{Corpus} & \textbf{Algorithm} & \textbf{Accuracy}(\%) & \textbf{Weighted F1-Score}(\%) \\ \hline
Original & RandomForest+BoW & 50.64 & 40.11 \\
 & SVM+BoW & 58.00 & 56.87 \\
 & LSTM+word2Vec & 53.39 & 53.30 \\
 & LSTM+fastText & 54.25 & 53.77 \\
 & \textbf{CNN+word2Vec} & \textbf{59.74} & \textbf{59.74} \\
 & CNN+fastText & 56.85 & 56.79 \\ \hline
Without Other & RandomForest+BoW & 50.64 & 49.14 \\
label & SVM+BoW & 63.12 & 62.45 \\
 & LSTM+word2Vec & 61.70 & 61.09 \\
 & LSTM+fastText & 62.06 & 61.83 \\
 & \textbf{CNN+word2Vec} & \textbf{66.48} & \textbf{66.34} \\
 & CNN+fastText & 63.47 & 62.68 \\ \hline
\end{tabular}
\end{table}

Through this, we concluded that, when removing Other label, the weighted F1-score reaches higher results with the same methods. Firstly, it is because of the decrease in number of emotion labels in the UIT-VSMEC corpus from 7 to 6 (anger, enjoyment, surprise, sadness, fear and disgust). Secondly, sentences not affected by noise data from Other label gives better results. To conclude, Other label does affect the performance of these algorithms. This will be our focus in building data in the future. Apart from that, we evaluate the learning curves of the four models proposed with the original dataset (the seven labels dataset) in which Random Forest and SVM utilize BoW feature, CNN and LSTM utilize word2vec embeddings. To conduct this experiment, we keep the test and the validation sets while putting training set in stages from 2,000 with 500 sentence-jump until the end of the set (5,548 sentences).

\subsection{Error Analysis}

\begin{figure}[htb]
\centering
\includegraphics[width=7cm]{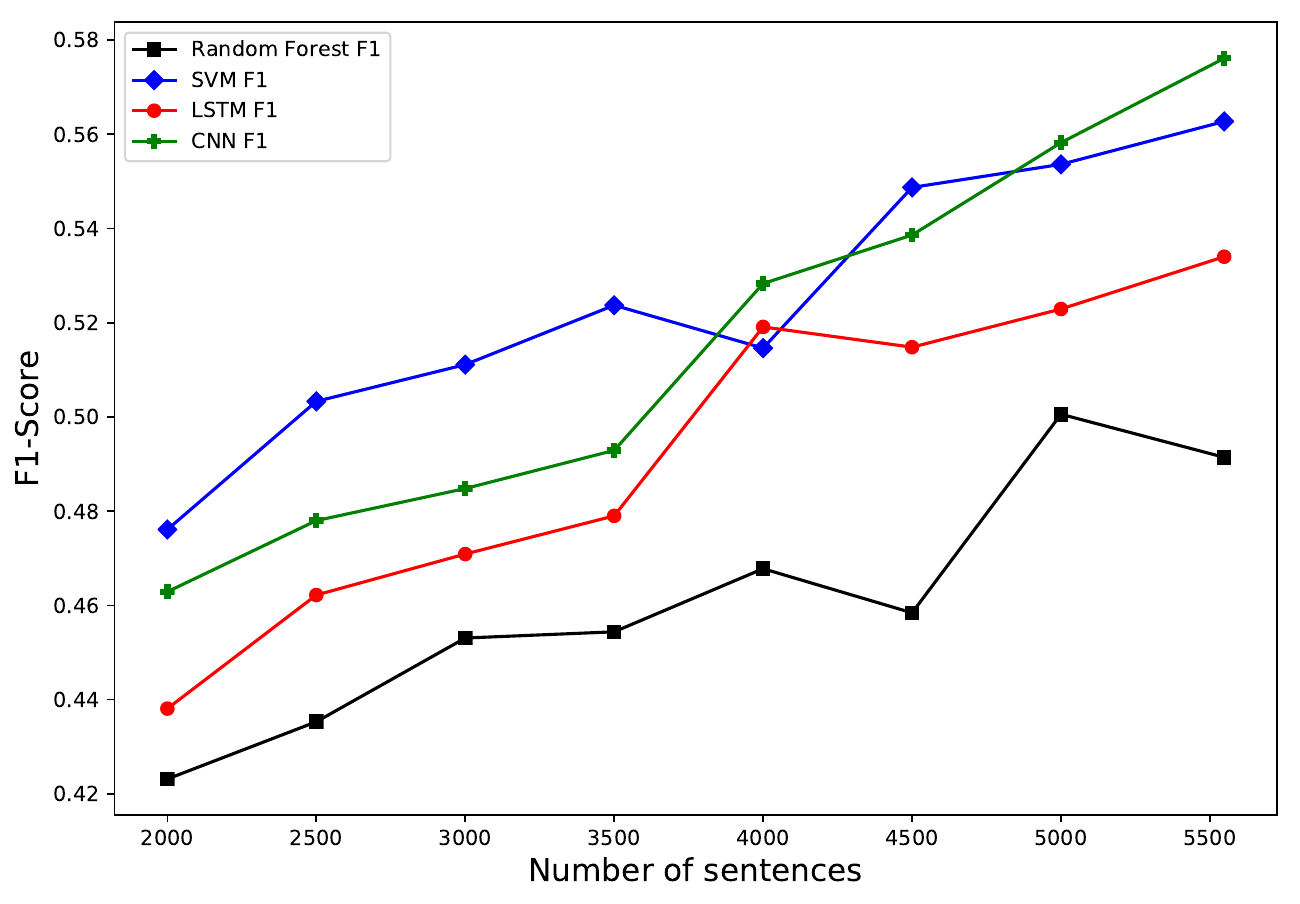}
\caption{Learning curves of the classification models on the UIT-VSMEC corpus.}
\label{fig:3}
\end{figure}

As can be seen in Figure \ref{fig:3}, when the size of the training set increases, so does the weighted F1-scores of the four models despite the slightly drop of 0.092\% in Random Forest when the number of the set grows from 5,000 to 5,500 sentences. In the meanwhile, compared to LSTM, the two deep learning models reach significant higher results, principally CNN combined with word2vec. Thus, we take this point to continue expanding the corpus as well as improving the performance of these models.

\begin{figure}[htb]
\centering
\includegraphics[width=8cm]{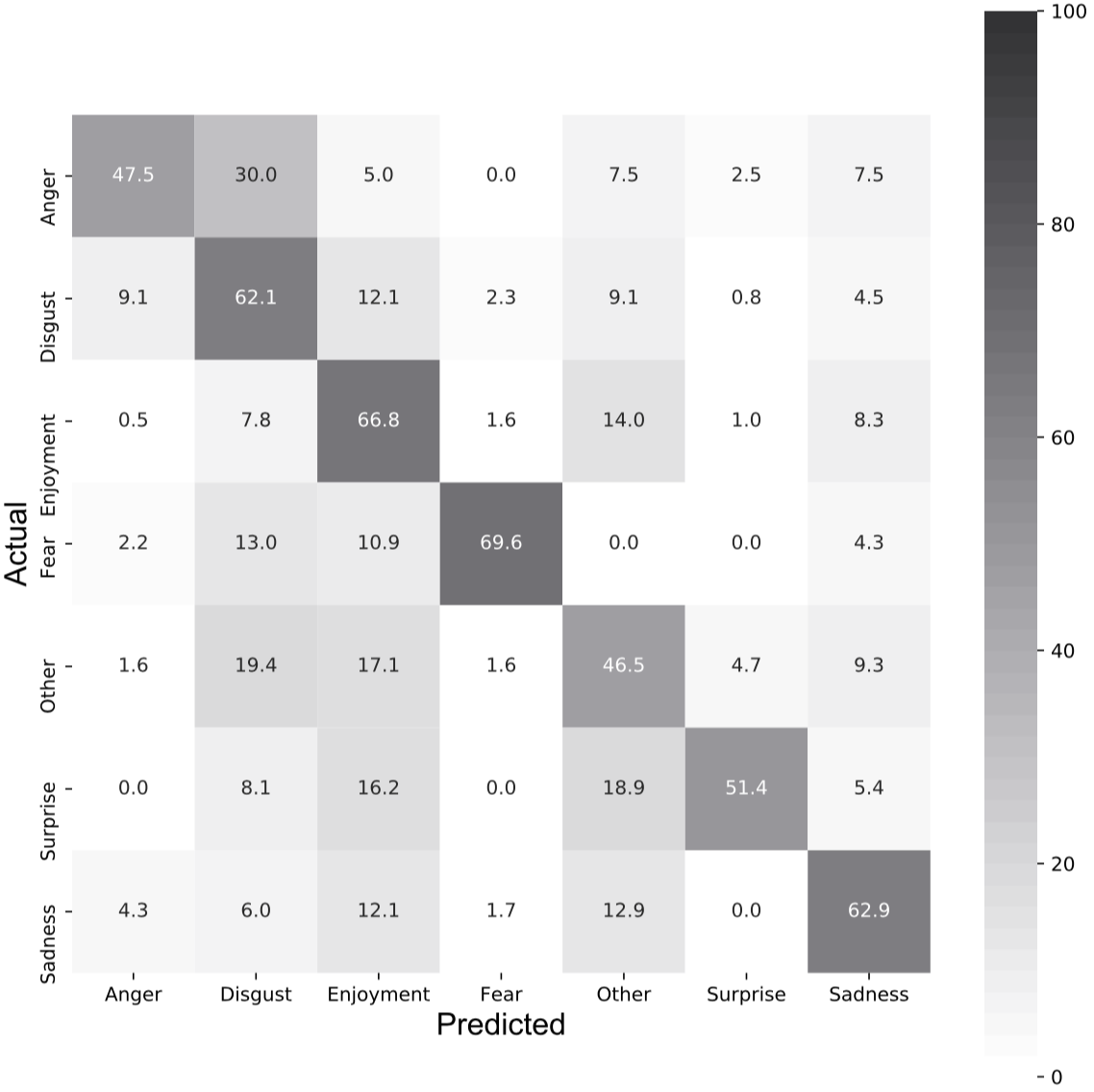}
\caption{Confusion matrix of the best classification model on the UIT-VSMEC corpus.}
\label{fig3}
\end{figure}

To demonstrate the performance of classification models, we use confusion matrix to visualize the ambiguity between actual labels and predicted labels.
Figure \ref{fig3} is the confusion matrix of the best classification model (CNN + word2vec) on the UIT-VSMEC corpus. As can be seen, the model performs well on classifying enjoyment, fear and sadness labels while it confuses between anger and disgust labels as their ambiguities are at the highest percentage of 39.1\%. There are two reasons causing this confusion. Primarily, it is the inherent vagueness of the definitions of anger and disgust construed through  \cite{Facial2007}. Secondly, the data limitation of these labels is an interference for the model to execute at its best. We noted this in our next step to continue building a thorough corpus for the task.
\section{Conclusion and Future Work} \label{conclusion}
In this study, we built a human-annotated corpus for emotion recognition for Vietnamese social media text for research purpose and achieved 6,927 sentences annotated with one of the seven emotion labels namely enjoyment, sadness, anger, surprise, fear, disgust and other with the annotation agreement of over 82\%. We also presented machine learning and deep neural network models used for classifying emotions of Vietnamese social media text. In addition, we reached the best overall weighted F1-score of 59.74\% on the original UIT-VSMEC corpus with CNN using the word2vec word embeddings.

In the future, we want to improve the quantity as well as the quality of the corpus due to its limitation of comments expressing emotions of anger, fear and surprise. Besides, we aim to conduct experiments using other machine learning models with distinctive features as well as deep learning models with various word representations or combine both methods on this corpus.

\section*{Acknowledgment}
We would like to give our thanks to the NLP@UIT research group and the Citynow-UIT Laboratory of the University of Information Technology - Vietnam National University Ho Chi Minh City for their supports with pragmatic and inspiring advice.

\bibliographystyle{splncs04}

\begin{thebibliography}{10}
\providecommand{\url}[1]{\texttt{#1}}
\providecommand{\urlprefix}{URL }
\providecommand{\doi}[1]{https://doi.org/#1}

\bibitem{PlabanKumarBhowmick}
{Bhowmick}, P.K., {Basu}, A., {Mitra}, P.: {An Agreement Measure for
  Determining Inter-Annotator Reliability of Human Judgements on Affective
  Tex}. In: {Proceedings of the workshop on Human Judgements in Computational
  Linguistics}. pp. 58--65. COLING 2008, Manchester, United Kingdom (2008)

\bibitem{Jointstockcompany}
company, J.S.: {The habit of using social networks of Vietnamese people 2018}.
  brands vietnam, Ho Chi Minh City, Vietnam (2018)

\bibitem{Ekman1993}
{Ekman}, P.: In: {Facial expression and emotion}. vol.~48, pp. 384--392.
  {American Psychologist} (1993)

\bibitem{Ekman}
{Ekman}, P.: In: {Emotions revealed: Recognizing faces and feelings to improve
  communication and emotional life}. p.~2007. {Macmillan} (2012)

\bibitem{PaulEkman}
{Ekman}, P., {Ekman}, E., {Lama}, D.: In: {The Ekmans' Atlas of Emotion} (2018)

\bibitem{Kim}
{Kim}, Y.: {Convolutional Neural Networks for Sentence Classifications}. In:
  {Proceedings of the 2014 Conference on Empirical Methods in Natural Language
  Processing (EMNLP)}. pp. 1746--1751. { Association for Computational
  Linguistics}, Doha, Qatar (2014)

\bibitem{Kiritchenko}
{Kiritchenko}, S., {Mohammad}, S.: {Using Hashtags to Capture Fine Emotion
  Categories from Tweets}. In: {Computational Intelligence}. pp. 301--326
  (2015)

\bibitem{RomanKlinger}
{Klinger}, R., {Clerc}, O.D., {Mohammad}, S.M., {Balahur}, A.: {IEST:WASSA-2018
  Implicit Emotions Shared Task}. pp. 31--42. 2017 AFNLP, Brussels, Belgium
  (2018)

\bibitem{BernhardKratzwald}
{Kratzwald}, B., {Ilic}, S., {Kraus}, M., S.~{Feuerriegel}, H.P.: {Decision
  support with text-based emotion recognition: Deep learning for affective
  computing}. pp. 24 -- 35. {Decision Support Systems} (2018)

\bibitem{SaifMohammad2017}
{Mohammad}, S., {Bravo-Marquez}, F.: {Emotion Intensities in Tweets}. In:
  {Proceedings of the Sixth Joint Conference on Lexical and Computational
  Semantics (*SEM)}. pp. 65--77. Association for Computational Linguistics,
  Vancouver, Canada (2017)

\bibitem{Mohammad}
{Mohammad}, S.M.: {\#Emotional Tweets}. In: {First Joint Conference on Lexical
  and Computational Semantics (*SEM)}. pp. 246--255. {Association for
  Computational Linguistics}, Montreal, Canada (2012)

\bibitem{Mohammad2018}
{Mohammad}, S.M., {Bravo-Marquez}, F., {Salameh}, M., {Kiritchenko}, S.:
  {SemEval-2018 task 1: Affect in tweets}. pp. 1--17. Proceedings of
  International Workshop on Semantic Evaluation, New Orleans, Louisiana (2018)

\bibitem{SaifMohammad}
{Mohammad}, S.M., {Xiaodan}, Z., {Kiritchenko}, S., {Martin}, J.: {Sentiment,
  emotion, purpose, and style in electoral tweets}. pp. 480--499. Information
  Processing and Management: an International Journal (2015)

\bibitem{Nguyen}
Nguyen: {Vietnam has the 7th largest number of Facebook users in the world}.
  Dan Tri newspaper (2018)

\bibitem{VLSPX}
{Nguyen}, H.T.M., {Nguyen}, H.V., {Ngo}, Q.T., {Vu}, L.X., {Tran}, V.M., {Ngo},
  B.X., {Le}, C.A.: {VLSP Shared Task: Sentiment Analysis}. In: {Journal of
  Computer Science and Cybernetics}. pp. 295--310 (2018)
  

\bibitem{KietVanNguyen}
{Nguyen}, K.V., {Nguyen}, V.D., {Nguyen}, P., {Truong}, T., {Nguyen}, N.L.T.:
  {UIT-VSFC: Vietnamese Students’ Feedback Corpus for Sentiment Analysis}.
  In: {2018 10th International Conference on Knowledge and Systems Engineering
  (KSE)}. pp. 19--24. {IEEE}, Ho Chi Minh City, Vietnam (2018)

\bibitem{PhuNguyen}
{Nguyen}, P.X.V., {Truong}, T.T.H., {Nguyen}, K.V., {Nguyen}, N.L.T.: {Deep
  Learning versus Traditional Classifiers on Vietnamese Students' Feedback
  Corpus}. In: {2018 5th NAFOSTED Conference on Information and Computer
  Science (NICS)}. pp. 75--80. Ho Chi Minh City, Vietnam (2018)

\bibitem{VuDucNguyen}
{Nguyen}, V.D., {Nguyen}, K.V., {Nguyen}, N.L.T.: {Variants of Long Short-Term
  Memory for Sentiment Analysis on Vietnamese Students’ Feedback Corpus}. In:
  {2018 10th International Conference on Knowledge and Systems Engineering
  (KSE)}. pp. 306--311. IEEE, Ho Chi Minh City, Vietnam (2018)

\bibitem{AurelienGeron}
Pedregosa, F., Varoquaux, G., Gramfort, A., Michel, V., Thirion, B., Grisel,
  O., Blondel, M., Prettenhofer, P., Weiss, R., Dubourg, V., Vanderplas, J.,
  Passos, A., Cournapeau, D., Brucher, M., Perrot, M., Duchesnay, E.:
  Scikit-learn: Machine learning in python. Journal of Machine Learning
  Research  \textbf{12},  2825--2830 (2011)

\bibitem{CarloStrapparava}
{Strapparava}, C., {Mihalcea}, R.: {SemEval-2007 Task 14: Affective Text}. In:
  {Proceedings of the 4th International Workshop on Semantic Evaluations
  (SemEval-2007)}. pp. 70--74. { Association for Computational Linguistics},
  Prague (2007)

\bibitem{TingweiWang}
{T. {Wang} and X. {Yang} and C. {Ouyang}}: {A Multi-emotion Classification
  Method Based on BLSTM-MC in Code-Switching Text: 7th CCF International
  Conference, NLPCC 2018, Hohhot, China, August 26–30, 2018, Proceedings,
  Part II.} pp. 190--199. Natural Language Processing and Chinese Computing
  (2018)

\bibitem{ZhongqingWang}
{Wang}, Z., {Li}, S.: {Overview of NLPCC 2018 Shared Task 1: Emotion Detection
  in Code-Switching Text: 7th CCF International Conference, NLPCC 2018, Hohhot,
  China, August 26–30, 2018, Proceedings, Part II}. pp. 429--433. Natural
  Language Processing and Chinese Computing (2018)

\bibitem{Facial2007}
{Zhang}, S., {Wu}, Z., {Meng}, H., {Cai}, L.: Facial expression synthesis using
  pad emotional parameters for a chinese expressive avatar. vol.~4738, pp.
  24--35 (09 2007)

\bibitem{YingjieZhang}
{Zhang}, Y., {Wallace}, B.C.: {A Sensitivity Analysis of (and Practitioners’
  Guide to Convolutional}. pp. 253--263. 2017 AFNLP, Taipei, Taiwan (2017)

\bibitem
{Nguyen}, K.V., {Nguyen}, N.L.T., 2016, October. {Vietnamese transition-based dependency parsing with supertag features}. In 2016 Eighth International Conference on Knowledge and Systems Engineering (KSE) (pp. 175-180). IEEE.

\bibitem{nguyen2014treebank}
Nguyen, D.Q., Pham, S.B., Nguyen, P.T., Le Nguyen, M., et al.: From treebankconversion to automatic dependency parsing for vietnamese. In: International Con-ference on Applications of Natural Language to Data Bases/Information Systems.pp. 196–207. Springer (2014)

\bibitem{nguyen2016vietnamese}
Nguyen, K.V., Nguyen, N.L.T.: Vietnamese transition-based dependency parsingwith supertag features. In: 2016 Eighth International Conference on Knowledgeand Systems Engineering (KSE). pp. 175–180. IEEE (2016)


\bibitem{bach2018empirical}
Bach, N.X., Linh, N.D., Phuong, T.M.: An empirical study on pos tagging forvietnamese social media text. Computer Speech \& Language50, 1–15 (2018)

\bibitem{nguyen2017word}
Nguyen, D.Q., Vu, T., Nguyen, D.Q., Dras, M., Johnson, M.: From word segmen-tation to pos tagging for vietnamese. arXiv preprint arXiv:1711.04951 (2017)

\bibitem{thao2007named}
Thao, P.T.X., Tri, T.Q., Dien, D., Collier, N.: Named entity recognition in viet-namese using classifier voting. ACM Transactions on Asian Language InformationProcessing (TALIP)6(4), 1–18 (2007)

\bibitem{nguyen2016approach}
Nguyen, L.H., Dinh, D., Tran, P.: An approach to construct a named entity anno-tated english-vietnamese bilingual corpus. ACM Transactions on Asian and Low-Resource Language Information Processing (TALLIP)16(2), 1–17 (2016)


\bibitem{Nguyen_2009}
Nguyen, D.Q., Nguyen, D.Q., Pham, S.B.: A vietnamese question answeringsystem. 2009 International Conference on Knowledge and Systems Engineering(Oct 2009). https://doi.org/10.1109/kse.2009.42,http://dx.doi.org/10.1109/KSE.2009.42

\bibitem{le2018factoid}
Le-Hong, P., Bui, D.T.: A factoid question answering system for vietnamese. In:Companion Proceedings of the The Web Conference 2018. pp. 1049–1055 (2018)







\end{thebibliography}

\end{document}